\documentclass[conference]{IEEEtran}
\IEEEoverridecommandlockouts
\usepackage{multirow}
\usepackage{boldline,multirow}
\usepackage{cite}
\usepackage{amsmath,amssymb,amsfonts}
\usepackage[noend]{algorithmic}
\usepackage{algorithm}
\usepackage{graphicx}
\usepackage{textcomp}
\usepackage{xcolor}
\usepackage{booktabs}
\usepackage{amsmath}
\usepackage{array} 
\usepackage{url}
\usepackage{hyperref}
\newcolumntype{C}[1]{>{\centering\arraybackslash}p{#1}}
\def\BibTeX{{\rm B\kern-.05em{\sc i\kern-.025em b}\kern-.08em
    T\kern-.1667em\lower.7ex\hbox{E}\kern-.125emX}}
\begin{document}

\title{
    Exploring the Limitations of Kolmogorov-Arnold Networks in Classification: Insights to Software Training and Hardware Implementation 
}

\author{
	\IEEEauthorblockN{Van Duy Tran\textsuperscript{1}, Tran Xuan Hieu Le\textsuperscript{2}, Thi Diem Tran\textsuperscript{2}, Hoai Luan Pham\textsuperscript{1}, Vu Trung Duong Le\textsuperscript{1}, Tuan Hai Vu\textsuperscript{1}, \\ Van Tinh Nguyen\textsuperscript{3}, and Yasuhiko Nakashima\textsuperscript{1}}
	\IEEEauthorblockA{
            \textsuperscript{1} Nara Institute of Science and Technology, 8916–5 Takayama-cho, Ikoma, Nara, 630-0192 Japan.\\
            \textsuperscript{2} University of Information Technology - VNUHCM, Vietnam.\\
            \textsuperscript{3} Le Quy Don technical University, Vietnam
.} 
}
\maketitle

\begin{abstract}
    Kolmogorov-Arnold Networks (KANs), a novel type of neural network, have recently gained popularity and attention due to the ability to substitute multi-layer perceptions (MLPs) in artificial intelligence (AI) with higher accuracy and interoperability. However, KAN assessment is still limited and cannot provide an in-depth analysis of a specific domain. Furthermore, no study has been conducted on the implementation of KANs in hardware design, which would directly demonstrate whether KANs are truly superior to MLPs in practical applications. As a result, in this paper, we focus on verifying KANs for classification issues, which are a common but significant topic in AI using four different types of datasets. Furthermore, the corresponding hardware implementation is considered using the Vitis high-level synthesis (HLS) tool. To the best of our knowledge, this is the first article to implement hardware for KAN. The results indicate that KANs cannot achieve more accuracy than MLPs in high complex datasets while utilizing substantially higher hardware resources. Therefore, MLP remains an effective approach for achieving accuracy and efficiency in software and hardware implementation.
\end{abstract}


\section{Introduction} \label{sec:introduction}
    Nowadays, Artificial Intelligence (AI) increasingly shows the advancement to solve problems in human life \cite{elbasi2022artificial}. In AI models, multi-layer perceptrons (MLPs) or fully connected neural networks provide the learnable ability and work as the most important component to approximate nonlinear functions \cite{hornik1989multilayer}. However, Kolmogorov–Arnold Networks (KANs) \cite{liu2024kan}, a new rise neural network architecture for AI and science, has recently been developed based on the Kolmogorov-Arnold representation theorem \cite{kantheory1, kantheory2} to replace MLPs due to its advantages in terms of accuracy and interpretability. Indeed, KANs in \cite{liu2024kan} have demonstrated their power by applying to issues such as partial differential equations (PDEs) \cite{evans2022partial}, continual learning \cite{wang2024comprehensive}, Knot theory \cite{manturov2018knot}, Anderson localization \cite{anderson1958absence}, etc.
    
    Although KAN could be applied in many applications, its endorsement is still not sufficient due to the limited number and complexity of possible scenarios. Hence, there is a need to implement more evaluations to fairly assess KANs' advantages and disadvantages compared to MLPs. Besides, there is no evaluation of KANs on hardware design which is essential to ensure KANs are acceptable for applying in practical applications efficiently and cost-effectively. Therefore, it is necessary to analyze KANs in terms of hardware design to clarify any potential concerns regarding the trade-off between accuracy and utilization of hardware resources.
    
    Though KANs have just recently been developed and published, there have been many works applying KANs to a wide range of applications. In \cite{bozorgasl2024wav},  Bozorgasl and Chen have developed Wavelet Kolmogorov-Arnold Networks to improve the original KANs for Wavelet transform and achieved better accuracy and shorter training time compared to MLPs. In addition, the authors in \cite{vaca2024kolmogorov} demonstrate the use of KANs in analyzing time series data for a satellite traffic forecasting task. Their findings indicate that KANs outperform MLPs in terms of lower error metrics, higher accuracy, and a reduced number of learnable parameters. Furthermore, KANs are also used to solve PDEs and gain outstanding accuracy in comparison with MLPs in \cite{wang2024kolmogorov} except for the complex geometry issue. Moreover, the authors in \cite{kundu2024kanqas} used KANs in quantum architecture search (QAS) and obtained notable results although the execution time of KANs-based systems is higher than MLPs-based systems. Additionally, the works in \cite{bodner2024convolutional, cheon2024kolmogorov} used KANs for the classification problems which are a common but important problem in MLPs-based models and achieve good results compared to MLPs. However, the experimental results are still limited to a small range of classification problems and cannot ensure the applicability of KANs in other classification issues. Besides, all the above previous works only implement their ideas on software platforms to obtain accuracy and there is no evaluation of hardware design about the trade-off between accuracy and hardware resource usage. 
    
    Therefore, in this paper, we introduce an analysis of four kinds of different classification problems to evaluate the effectiveness of KANs on classification problems. These classification problems consist of moons binary classification \cite{scikit-learn-make_moons}, three-label wine classification \cite{misc_wine_109}, seven-label dry bean classification \cite{misc_dry_bean_602}, and mushroom binary classification \cite{mushroom-dataset}. In addition, we also execute the hardware designs of the MLPs and KANs from the four datasets given above to assess the hardware's efficiency and illuminate the relationship between accuracy and hardware resources.
    
    The subsequent sections of this paper are organized as follows: Section \ref{sec:background_knowledge} provides the background knowledge about KANs. Section \ref{sec:proposed_methods} provides an in-depth description of the proposed approaches, which includes information on software training and hardware implementation. In Section \ref{sec:results_and_discussion}, we extensively assess and compare MLPs and KANs. Finally, Section \ref{sec:conclusion} provides a summary of this paper. Codes are available at: \url{github.com/Zeusss9/KAN_Analysis}

\section{Background Knowledge} \label{sec:background_knowledge}
    Unlike MLPs based on the universal approximation theorem, KANs, a new promising architecture, focus on the Kolmogorov-Arnold representation theorem to approximate nonlinear functions. Then, the Kolmogorov-Arnold representation theorem and the way KANs are implemented based on this theorem will be presented in this section.

    \subsection{Kolmogorov-Arnold Representation Theorem}
         The Kolmogorov-Arnold representation theorem, also known as the superposition theorem, was established by Vladimir Arnold and Andrey Kolmogorov. It states that any multivariate continuous function $f(x_1, \ldots, x_n)$ on a bounded domain can be represented as a finite composition of continuous functions of a single variable and the binary operation of addition. 
        
        \begin{equation} 
            f(x_1, \ldots, x_n) = \sum_{q=1}^{2n+1} \Phi_q \left( \sum_{p=1}^{n} \phi_{q,p}(x_p) \right),
            \label{eq:original_KAN}
        \end{equation} 
        
        where $\phi_{q,p} : [0,1] \to \mathbb{R}$ and $\Phi_q : \mathbb{R} \to \mathbb{R}$. This formula shows that any multivariate function can fundamentally be presented as an appropriate combination of univariate (1D) functions.
        
    \subsection{Kolmogorov-Arnold Network Architecture}
\begin{figure} 
            \centering
    	\includegraphics[width=0.5\textwidth]{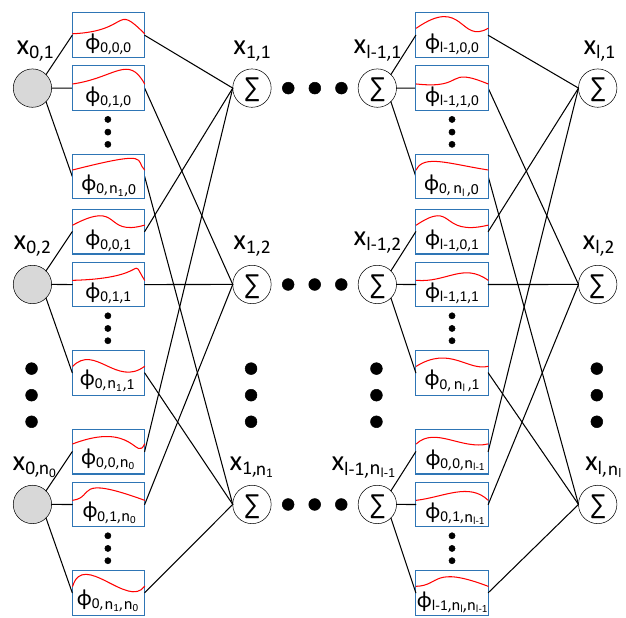}
    	\caption{The generalized KAN architecture}
            \label{fig:generalized_KAN_arch}
        \end{figure}

        KAN is a neural network \cite{liu2024kan} developed from the Kolmogorov-Arnold representation theorem with a huge improvement compared to previous existing works associated with KAN. In detail, each 1D function can be parameterized as a B-spline curve with learnable coefficients of local B-spline basis functions. This is because all of the functions that need to be learned in Eq. \ref{eq:original_KAN} are univariate functions. In addition, KAN utilizes the MLP architecture to construct a wider and deeper KAN instead of common networks of size ($n$, $2n+1$,1). A KAN architecture consists of KAN layers stacked to form a complete network. Then, a KAN layer with $n_{in}$-dimensional inputs and $n_{out}$-dimensional outputs can be described as follows in a matrix of 1D functions:
        
        \begin{equation} 
            \Phi = \{ \phi_{q,p} \}, \quad p = 1, 2, \cdots, n_{\text{in}}, \quad q = 1, 2, \cdots, n_{\text{out}}
            \label{eq:KAN_layer}
        \end{equation} 

        where $\phi_{q,p}$ functions are spline functions consisting of trainable parameters. Consequently, the Kolmogorov-Arnold representations in Eq. (\ref{eq:original_KAN}) are compositions of two such KAN layers that can be made wider and deeper. Then, a generalized KAN model shown in Fig. \ref{fig:generalized_KAN_arch} can be clearly stated as $[n_0, n_1, \cdots, n_L]$, with the number of nodes in each layer respectively. Please note that $L$ is the number of layers in a KAN architecture except the input layer.

        In this KAN architecture, there are $\phi_{l,j,i}(x_{l,i})$ functions between the neuron $i^\text{th}$ of the layer $l^\text{th}$ and the neuron $j^\text{th}$ of the layer $(l+1)^\text{th}$. They replace the role of weight in a typical MLP model and become trainable parameters as mentioned above. In general, a $\phi()$ can be approximated using B-splines (a type of spline), which are smooth curves consisting of control points. In more detail, splines are mathematical functions employed to smoothly and continuously interpolate or approximate data points. A spline is determined by two parameters: k, which represents the degree of the polynomial functions used for interpolation or approximation, and G, which represents the number of subintervals between adjacent control points. Using subintervals, spline interpolation joins the data points to produce a smooth curve made up of G + 1 grid points.

        Besides, the activation function at the neuron $j^\text{th}$ in the layer $(l+1)^\text{th}$ in a normal MLP model also is transformed into the summation of the $\phi_{l,j,i}(x_{l,i})$ results and is expressed as follows:

        \begin{equation}
            x_{l+1,j} = \sum_{i=1}^{n_l} \phi_{l,j,i}(x_{l,i}), \quad j = 1, \cdots, n_{l+1},
            \label{eq:activation_function}
        \end{equation} 
        
        On the other way, this can be expressed in a matrix form as follows: 
        
        \begin{equation}
            \mathbf{x}_{l+1} = 
            \underbrace{
            \left( 
            \begin{array}{cccc}
                \phi_{l,1,1}(\cdot) & \phi_{l,1,2}(\cdot) & \cdots & \phi_{l,1,n_l}(\cdot) \\
                \phi_{l,2,1}(\cdot) & \phi_{l,2,2}(\cdot) & \cdots & \phi_{l,2,n_l}(\cdot) \\
                \vdots & \vdots & \ddots & \vdots \\
                \phi_{l,n_{l+1},1}(\cdot) & \phi_{l,n_{l+1},2}(\cdot) & \cdots & \phi_{l,n_{l+1},n_l}(\cdot)
            \end{array} 
            \right)
            }_{\Phi_l} \mathbf{x}_l,
        \end{equation} 
        
        where ${\Phi_l}$ is the function matrix according to the $l^\text{th}$ KAN layer. Hence, a common KAN model can be made from $L$ KAN layers with an input vector $x^0 \in \mathbb{R}^{n_0}$:
        
        \begin{equation}
            \text{KAN}(\mathbf{x}) = (\Phi_{L-1} \circ \Phi_{L-2} \circ \cdots \circ \Phi_1 \circ \Phi_0) \mathbf{x}.
        \end{equation}  

        Finally, the ${\phi()}$ function in each node in the hidden layers and output layer will be approximated by a symbolic function. The number of nodes at the output layer will correspond to the number of final symbolic formulas formed there. Then, the symbolic formulas will be utilized to predict data, with input data serving as the variables for each symbolic formula.

\section{Proposed methods} \label{sec:proposed_methods}
    In this section, we present in detail the software training method and hardware implementation approach of the four kinds of classification datasets as mentioned in Section \ref{sec:introduction}. 

    \subsection{Software training method} \label{subsec:software_method}
        Ensuring the diversity, complexity, and reality of testing datasets is critical for proving the ability of a kind of neural network. Therefore, in this section, we proceed to analyze KANs and MLPs on four different datasets with increasing complexity and reality as follows:
        
        \subsubsection{Moons Binary Classification}
            The Moons dataset is a synthetic dataset created by generating any required amount of data (10000 generated data samples in this case) using the \textit{make\_moons} of the \textit{scikit-learn}. It includes two intersecting semicircles making binary classification with two features for each input. In addition, the complexity of the Moons dataset is relatively low, as it has a distinct non-linear boundary that separates the two classes. Therefore, it is completely appropriate for evaluating the fundamental abilities and initial efficiency of neural networks, specifically KAN in this case. With the MLP model, the numbers of neurons in each layer are 2, 8, 4, and 2 respectively with the ReLU activation function in hidden layers and the Sigmoid activation function in the output layer. On the other hand, the KAN model is configured with $grid=3$, $k=3$, and the model size is (2,2,1).

        \subsubsection{Three-Label Wine Classification}
            The Wine dataset is a dataset about the Wine quality that comprises three labels and 13 features resulting from a chemical analysis. Hence, the Wine dataset has a moderate level of complexity compared to the Moons dataset. Nevertheless, the number of patterns in the Wine dataset is relatively low (178) even if we use the \textit{smote} method to increase the number of samples (213). Nonetheless, the Wine dataset is sufficient to assess the KAN's efficiency at this level of complexity. For evaluation, the MLP model consists of four layers with 13, 32, 8, and 3 neurons correspondingly, with ReLU function at the hidden layers and Softmax at the output layer. Meanwhile, the KAN model has been configured with $grid=3$ and $k=3$ and the model size is (13,4,1).
            
        \subsubsection{Seven-Label Dry Bean Classification}
            The Dry Bean dataset is more complex than the previous two datasets since it consists of seven classes and 16 features on each data sample. Besides, it has 13,611 data pattern which is sufficient for assessing KANs by requiring models to effectively handle a higher number of features and more diverse class distributions. Therefore, it can determine whether the model effectively applies to more intricate situations in the real world. During the training process, the MLP model consists of five layers with 16, 20, 15, 10, and 7 neurons accordingly, with ReLU acting as an activation function in hidden levels and Softmax acting as an activation function at the output layer. Besides, the KAN model has $grid=6$, $k=3$, and a model size of (16,2,7).
            
        \subsubsection{Mushroom Binary Classification}
            The Mushroom dataset is also a practical dataset consisting of 8,124 instances of mushrooms. Each data sample in the dataset is characterized by eight features and categorized into one of two categories: edible or poisonous. Although the Mushroom dataset is used for binary classification tasks, its complexity comes from a variety of practical features and the enormous amount of data samples. Therefore, it can be utilized to validate the performance of KANs compared to MLPs. For evaluation, the MLP model has 16, 20, 15, 10, and 7 neurons in each layer, with ReLU in the hidden layers and Softmax in the output layer. Besides, the KAN model has grid=6, k=3, and a model size of (8,24,2).
        \begin{figure} 
            \centering
            \includegraphics[width=0.48\textwidth]{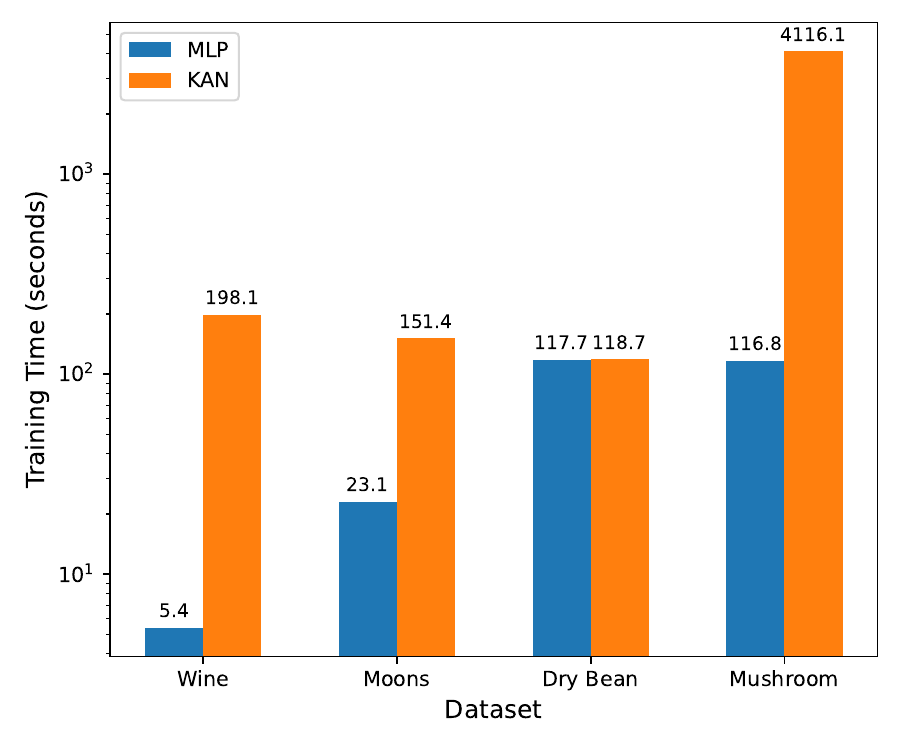}
            \caption{The training time between MLPs and KANs}
            \label{fig:training_time}
        \end{figure}

                 \begin{figure*}
            \centering
            \includegraphics[width=1\textwidth]{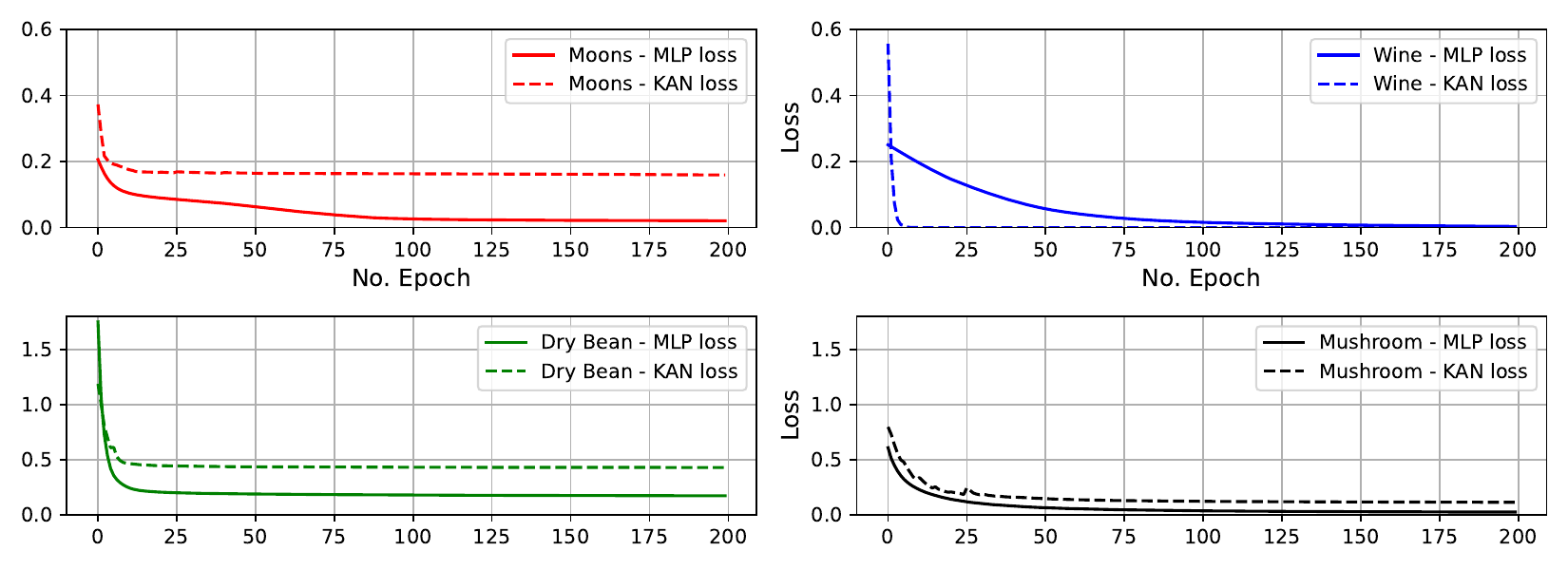}
            \caption{The loss during training processes of MLPs and KANs}
            \label{fig:loss}
        \end{figure*}
        
    \subsection{Hardware implementation approach}
        Implementing an algorithm or a software program into a hardware design manually always takes a lot of effort and time. Hence, High-level synthesis (HLS) appears to transform high-level algorithmic descriptions written in languages such as C/C++ into hardware description languages (HDL). HLS has the advantages of reducing development time, simplifying design modifications, and facilitating system maintenance compared to manual hardware description language (HDL) coding. Through the automation of mapping intricate C/C++ programs to hardware, High-Level Synthesis (HLS) provides the capability to quickly implement various optimization strategies to achieve superior performance and efficiency. Hence, HLS becomes a remarkable methodology for implementing hardware design.

        Vitis HLS \cite{vitishls}, one of the best HLS software tools, is developed by Xilinx which allows the design using high-level programming languages such as C/C++, instead of traditional HDL such as Verilog and Very High-Speed Integrated Circuit Hardware Description Language (VHDL). Hence, the generated HDL from Vitis HLS can be directly implemented on field-programmable gate array (FPGA) devices. Furthermore, it provides a robust development environment that shortens the design cycle and increases productivity. It also includes the Vitis HLS flow and tools for simulating, synthesizing, modifying, and optimizing logic circuits to achieve desired architecture faster than writing HDL code manually. Therefore, Vitis HLS is a great option for rapidly and effectively implementing hardware.

\section{Experimental setup} \label{sec:experimental_setup} 
    To evaluate the performance of KANs and MLPs in the software training process, the four datasets mentioned in Section \ref{subsec:software_method} are utilized to evaluate the capability of KANs when dealing with various types of complicated datasets. Furthermore, the training processes are conducted on a Windows 11 computer with the NVIDIA RTX 3060 graphics processing unit (GPU). 
    
    In addition, we utilized the Vitis HLS 2023.1 tool to synthesize KANs and MLPs in C/C++ programs throughout the hardware implementation process. In detail, the symbolic formulas after symbolic transformations in Python will be converted manually to C/C++ while we use Py2C to convert MLPs in Python to C/C++ programs. It can ensure the preservation of correctness when compared to software accuracy values, while also achieving the hardware design quickly and effectively. To offer a fair comparison between KANs and MLPs, all hardware implementations of both KANs and MLPs are set up according to the following configuration: 

        \begin{table*}[t]
        \small
        \centering
        \renewcommand{\arraystretch}{1.4}
        \caption{Software training results between MLPs and KANs}
        \label{tab:software_results}
            \begin{tabular}{|C{1.4cm}|C{0.8cm}|C{1.7cm}|C{1.1cm}|C{0.9cm}|C{0.7cm}|C{0.7cm}|C{0.9cm}|C{0.9cm}|C{0.7cm}|C{0.7cm}|C{1cm}|C{1cm}|}
            \hline
            \multirow{2}{*}{Dataset} & Model & Model & Spline & No. & \multicolumn{2}{c|}{Accuracy} & \multicolumn{2}{c|}{ET** (s)} & \multicolumn{2}{c|}{Power (W)} &\multicolumn{2}{c|}{PDP{\textdagger} (W*s)}\\
            \cline{6-13}
             & Type & Size & Info & Params & Pre* & Post$\ulcorner$ & Pre* & Post$\ulcorner$ & Pre* & Post$\ulcorner$ & Pre* & Post$\ulcorner$\\
            \hline
            \multirow{2}{*}{\textbf{Moons}} & MLP & 2,8,4,1 & N/A & 65 & \multicolumn{2}{c|}{0.968} & \multicolumn{2}{c|}{0.262} & \multicolumn{2}{c|}{19.52} & \multicolumn{2}{c|}{5.114}\\
            \cline{2-13}
            & KAN & 2,2,1 & G=3/k=3 & 36 & 0.969 & 0.968 & 0.32 & 95.942 & 20.41 & 21.48 & 6.531 & 2060.83 \\ 
            \hline
            \multirow{2}{*}{\textbf{Wine}} & MLP & 13,32,8,3 & N/A & 739 & \multicolumn{2}{c|}{0.984} & \multicolumn{2}{c|}{0.089} & \multicolumn{2}{c|}{20.38} & \multicolumn{2}{c|}{1.813}\\
            \cline{2-13}
            & KAN & 13,4,3 & G=3/k=4 & 448 & 0.984 & 0.973 & 0.081	& 1.696 & 21.48 & 22.41 & 1.739 & 38.007 \\
            \hline
            \multirow{2}{*}{\textbf{Dry Bean}} & MLP & 16,20,15,10,7 & N/A & 892 & \multicolumn{2}{c|}{0.921} & \multicolumn{2}{c|}{0.37} & \multicolumn{2}{c|}{21.33} & \multicolumn{2}{c|}{7.892}\\
            \cline{2-13}
            & KAN & 16,2,7 & G=6/k=3 & 414 & 0.924 & 0.919  & 0.3 &	520.56 & 22.78 & 21.81 & 6.834 & 11353.5 \\
            \hline
            \multirow{2}{*}{\textbf{Mushroom}} & MLP & 8,64,64,2 & N/A & 4866 & \multicolumn{2}{c|}{0.99} & \multicolumn{2}{c|}{1.067} & \multicolumn{2}{c|}{21.09} & \multicolumn{2}{c|}{22.5}\\
            \cline{2-13}
            & KAN & 8,24,2 & G=6/k=3 & 2160 & 0.994 & 0.558  & 0.57 & 4501.3 & 43.96 & 21.03 & 25.057 & 94662.3 \\
            \hline

            \multicolumn{13}{l}{Pre*: KANs before transforming to the symbolic formula; Post$\ulcorner$: KANs after transforming to the symbolic formula.} \\
            \multicolumn{13}{l}{ET**: Execution time of a datasetl; PDP\textdagger: Power Delay Product = ET * Power.} \\
            \end{tabular}
        \end{table*}
        
    \begin{itemize}
        \item{Operational frequency: 100 MHz.}
        \item {Utilized FPGA for synthesis and implementation: a Xilinx Zynq Ultrascale+ ZCU104 field-programmable gate array (FPGA) board.}
        \item{Kind of number for data processing: Floating-point 32 bits.}
    \end{itemize}

\section{Results and discussion} \label{sec:results_and_discussion}
    This section presents and analyzes the results after software training and hardware post-implementation. Then, we can make clear the effectiveness of KANs compared to MLPs.

    \subsection{Software results} \label{subsec:software_results}
        To ensure a fair comparison, we train all KAN and MLP models over 200 epochs, using a learning rate of 0.0002. Furthermore, we employ the Adam optimizer for all MLP models, while the L-BFGS optimizer (the default optimizer of KANs) is utilized for KAN models.

        Fig. \ref{fig:training_time} exhibits the training time difference between MLPs and KANs for each of the four datasets. Except for the Dry Bean dataset's training time, all three of the other datasets always show that KANs require substantially longer training times than MLPs, ranging from \textbf{6.55} times (151.4 vs 23.1 s) to \textbf{36.68} times (198.1 vs 5.4 s). This is because of the trainable spline function in KANs as mentioned in \cite{liu2024kan}. Besides, Fig. \ref{fig:loss} presents the loss values over training processes. In particular, we employ the Cross-Entropy loss function for the Dry Bean and Mushroom datasets and the Mean Squared Error loss function for the Moons and Wine datasets. Except for the Wine dataset, MLPs continually have faster loss reduction and lower loss values than KANs. Overall, regarding training time and loss reduction, KANs are not better than MLPs.

        In addition, table \ref{tab:software_results} shows the training configuration and results. The Pre and Post values refer to the KANs' results before and after transforming to symbolic formulas. Additionally, the number of KAN parameters is calculated as (\ref{eq:KAN_param}):

        \begin{equation} \label{eq:KAN_param}
            \begin{aligned}
                \text{Total Parameters} = \left( \sum_{l=0}^{L-1} n_l \times n_{l+1} \right) \times (G + k)
            \end{aligned}
        \end{equation}

        where $L$ is the number of layers while $n_l$ is the number of neurons in the $l^{th}$ layer. In addition, each spline is of order $k$ and consists of $G$ intervals.
        
        In detail, the KAN models have approximately half the number of trainable parameters compared to the MLP models due to the small size of networks as mentioned in \cite{liu2024kan}. In addition, the Pre Accuracy demonstrates that KANs are not significantly better than MLPs in classification problems from the Moons to the Mushroom datasets (\textbf{0.969} vs. \textbf{0.968} / \textbf{0.984} vs. \textbf{0.984} / \textbf{0.921} vs. \textbf{0.924} / \textbf{0.99} vs. \textbf{0.994}). Moreover, the Post Accuracy reveals a decrease in accuracy values compared to the Pre Accuracy when symbolic functions are used instead of matrix multiplications after the training process. This phenomenon becomes more obvious when the complexity of the dataset rises from the Moons to the Mushroom datasets. Particularly, the data has a random shape, making it challenging for any single symbolic function to perfectly fit the data. Consequently, even with considerable effort to figure out the most appropriate single symbolic function or complicated symbolic formula, accuracy is unavoidably affected. Moreover, when the model size increases due to the complexity of the dataset, there is a corresponding increase in the loss of accuracy caused by the accumulation of losses across the whole network. Furthermore, the process of converting KANs to symbolic formulas takes significant effort from developers to adjust KAN models and minimize the loss of accuracy manually. Consequently, the training procedure is not only time-consuming but also requires a substantial amount of time to convert a KAN model into symbolic formulas. Besides, the Power Delay Product (PDP) shows that MLP is almost better than KANs in all cases with all Pre and Post values. It indicates that KANs consume much more power consumption over time in comparison with MLPs. 
        
        In general, KANs fail to demonstrate higher accuracy than MLPs, and the symbolic formula representation of KANs performs even worse than MLPs in classification challenges. Furthermore, KANs also need a substantial amount of time and effort from developers to create symbolic formulas in the final stages. Based on the above findings, MLPs remain an effective choice in comparison to KANs for classification tasks.
        
    \subsection{Hardware results} \label{subsec:hardware_results}
        \begin{table*}[t]
        \small
        \centering
        \renewcommand{\arraystretch}{1.4}
        \caption{Hardware post-implementation results between MLPs and KANs (Post)}
        \label{tab:hardware_results}
            \begin{tabular}{|C{1.7cm}|C{1.2cm}|C{1cm}|C{1.3cm}|C{1.3cm}|C{1.3cm}|C{1.3cm}|C{1.2cm}|C{1.6cm}|C{1.6cm}|}
            \hline
            \multirow{2}{*}{Dataset} & Model & Freq. & \multicolumn{4}{c|}{Hardware Resources} & Power & \multicolumn{2}{c|}{Latency} \\
            \cline{4-7} \cline{9-10}
             & Type & (MHz) & \#BRAMs & \#DSPs & \#FFs & \#LUTs & (W) & \#Cycles & Time (ns) \\
            \hline
            \multirow{2}{*}{\textbf{Moons}} & MLP & \multirow{8}{*}{100} & 0 & 17 & 3364 & 3809 & 0.663 & 149 & 1490 \\
            \cline{2-2}
            \cline{4-10}
            & KAN &  & 10 & 120 & 8622 & 17877 & 0.717 & 128 & 1280 \\
            \cline{1-2}
            \cline{4-10}
            \multirow{2}{*}{\textbf{Wine}} & MLP &  & 2 & 17 & 9936 & 6974 & 0.675 & 699 & 6990 \\
            \cline{2-2}
            \cline{4-10}
            & KAN &  & 132 & 950 & 74741 & 146843 & 1.349 & 688 & 6880 \\
            \cline{1-2}
            \cline{4-10}
            \multirow{2}{*}{\textbf{Dry Bean}} & MLP &  & 0 & 17 & 11328 & 8894 & 0.67 & 835 & 8350 \\
            \cline{2-2}
            \cline{4-10}
            & KAN &  & 781 & 9111 & 734544 & 1677558 & 14.802 & 1896 & 18960 \\
            \cline{1-2}
            \cline{4-10}
            \multirow{2}{*}{\textbf{Mushroom}} & MLP &  & 4 & 17 & 18903 & 10932 & 0.673 & 3128 & 31280 \\
            \cline{2-2}
            \cline{4-10}
            & KAN &  & 1347 & 16299 & 1337291 & 3112275 & - & 3434 & 34340 \\
            \hline

            \multicolumn{10}{l}{- : The power consumption cannot be aggregated since the hardware resources exceed all of our FPGA boards.} \\
            \end{tabular}
        \end{table*}
        
        The post-implementation results of the hardware are presented in Table \ref{tab:hardware_results} for comparing KANs (Post) and MLPs. It shows that the hardware resources of KANs are significantly greater than MLPs across all four kinds of datasets while table \ref{tab:software_results} shows that the parameters of KANs are lower than MLPs. These results indicate that implementing symbolic formulas on hardware requires much more hardware resources compared to normal matrix multiplication in MLPs. Moreover, when the size of KAN's models increases, there is a corresponding rise in hardware resources required. Indeed, the BRAMs used in KANs are higher than MLPs, ranging from \textbf{10} to \textbf{1347}, while the DSPs employed in KANs increase significantly, ranging from \textbf{7.06} times (\textbf{120} vs. \textbf{17}) to \textbf{958.76} times (\textbf{16299} vs. \textbf{17}) compared to MLPs. In addition, the FFs of KANs are significantly higher than those of MLPs, fluctuating between \textbf{2.56} times (\textbf{8622} vs. \textbf{3364} and \textbf{70.74} times (\textbf{1337291} vs. \textbf{18903}). On the other hand, the LUTs of MLPs are substantially less than those of KANs, varying from \textbf{4.7} times (\textbf{3809} vs. \textbf{17877} to \textbf{284.7} times (\textbf{10932} vs. 3112275). As a result, the power consumption of KAN is significant compared to MLP due to the high amount of resources used. Moreover, even the latency of KANs is not better than MLPs with the longer latency ranging from \textbf{1.1} times (3434 vs. 3128 cycles) to \textbf{2.27} times (\textbf{1896} vs. \textbf{835} cycles).
        
        Overall, MLPs outperform KANs in both performance and hardware resources. Based on that, MLPs remain a significantly superior option compared to KAN  when dealing with the classification problem on hardware.
        
\section{Conclusion} \label{sec:conclusion}
    In this study, we utilized four distinct datasets with various levels of complexity to conduct an extensive assessment of KANs against MLPs for classification tasks in both software training and hardware implementation. The results reveal that KANs, despite their theoretical advantages in accuracy and interpretability, do not outperform MLPs in practical classification tasks. In addition to their hardware implementation, KANs are less effective for practical applications since they use a lot of resources and have higher latency than MLPs. In conclusion, MLPs continue to be a more practical and resource-efficient option for classification issues, even though KANs have some theoretical advantages. In the future, we can improve KAN architectures in both software training and hardware implementation to facilitate KANs' performance as well as increase the number of evaluated datasets to gain more comprehensive analysis.
    
\section*{Acknowledgment}
This work was supported by JST-ALCA-Next Program Grant Number JPMJAN23F4, Japan. The research has been partly executed in response to the support of JSPS, KAKENHI Grant No. 22H00515, Japan.

% \section*{References}
\bibliographystyle{IEEEtran}
\bibliography{references.bib}

\end{document}